\documentclass[conference]{IEEEtran}
\IEEEoverridecommandlockouts
\usepackage{cite}
\usepackage{amsmath,amssymb,amsfonts}
\usepackage{algorithmic}
\usepackage{graphicx}
\usepackage{textcomp}
\usepackage{xcolor}

\usepackage[utf8]{inputenc} 
\usepackage[T1]{fontenc}    
\usepackage{hyperref}       
\usepackage{url}            
\usepackage{booktabs}       
\usepackage{amsfonts}       
\usepackage{nicefrac}       
\usepackage{microtype}      
\usepackage{lipsum}		
\usepackage{graphicx}
\usepackage{doi}
\usepackage{xcolor}
\usepackage{float}

\graphicspath{ {./} }

\def\BibTeX{{\rm B\kern-.05em{\sc i\kern-.025em b}\kern-.08em
    T\kern-.1667em\lower.7ex\hbox{E}\kern-.125emX}}
\begin{document}

\title{Lip reading using external viseme decoding}

\author{\IEEEauthorblockN{1\textsuperscript{st} Javad~Peymanfard}
\IEEEauthorblockA{\textit{School of Computer Engineering} \\
\textit{Iran University of Science and Technology}\\
Tehran, Iran \\
javad\_peymanfard@comp.iust.ac.ir}
\and
\IEEEauthorblockN{2\textsuperscript{st} Mohammad Reza Mohammadi}
\IEEEauthorblockA{\textit{School of Computer Engineering} \\
\textit{Iran University of Science and Technology}\\
Tehran, Iran \\
mrmohammadi@iust.ac.ir}
\and
\IEEEauthorblockN{3\textsuperscript{st} Hossein Zeinali}
\IEEEauthorblockA{\textit{Department of Computer Engineering} \\
\textit{Amirkabir University of Technology}\\
Tehran, Iran \\
hzeinali@aut.ac.ir}
\and
\IEEEauthorblockN{4\textsuperscript{st} Nasser Mozayani}
\IEEEauthorblockA{\textit{School of Computer Engineering} \\
\textit{Iran University of Science and Technology}\\
Tehran, Iran \\
mozayani@iust.ac.ir}
}
\maketitle

\begin{abstract}
Lip-reading is the operation of recognizing speech from lip movements. This is a difficult task because the movements of the lips when pronouncing the words are similar for some of them. Viseme is used to describe lip movements during a conversation. This paper aims to show how to use external text data (for viseme-to-character mapping) by dividing video-to-character into two stages, namely converting video to viseme, and then converting viseme to character by using separate models. Our proposed method improves word error rate by an absolute rate of 4\% compared to the normal sequence to sequence lip-reading model on the BBC-Oxford Lip Reading Sentences 2 (LRS2) dataset.
\end{abstract}


\section{Introduction}

Lip-reading is commonly used to understand human speech without hearing sound and by using only visual features. This ability is more common in people with hearing loss or hearing problems. Over the past years, several methods have been proposed for a person to lip-read \cite{matthews2002extraction}, but there is an important difference between these methods and the lip-reading methods suggested in AI. The purpose of the proposed methods for lip-reading by the machine is to convert visual information into words. This conversion takes place on two levels, which are described below. However, the main purpose of lip-reading by humans is to understand the meaning of speech and not to understand every single word of speech. Obviously, visemes are the main challenge in lip-reading. Visemes are the visual equivalent of phonemes \cite{chung2017lip}. In fact, each viseme refers to a group of phonemes in which the movement of the lips is the same, such as /b/, /p/, and /m/.

The traditional approaches of automatic lip-reading used hand-crafted features \cite{matthews2002extraction, zhou2011towards, potamianos2001improved, lan2009comparing}. Hidden Markov Model was also applied for modeling \cite{potamianos1998image, lan2010improving, morade2014novel}. However, with new large public datasets introduced in this area in recent years and deep learning methods used for this purpose, more appropriate methods have been proposed which are more accurate than even a professional lip reader. These datasets are divided into two groups, namely word-level and sentence-level lip-reading. In the first group, the lip-reading problem is a classification task. There are some vocabularies (classes) and the model aims at assigning an input video to a class. But in the second group, which is a sequence learning task, each sample (video) has a sentence-level text. This is a more challenging task than word-level prediction.

In this paper, we propose a method in which independent textual data from lip-reading datasets can be utilized to achieve higher accuracy for modeling. In this method, an external viseme decoding can be modeled using only textual data, so it is efficient to build it.

We employed a sequence-to-sequence model for viseme-to-character modeling to predict characters using a large amount of text data. A two-layer GRU with an attention mechanism is used in this model. The use of this external model has increased the accuracy of the entire lip-reading process.

The paper is organized as follows: In section 2, we review the most important related works. In this section, we also discuss the advantages and disadvantages of the available methods. In Section 3, we discuss the proposed model and describe it in detail. In Section 4, we discuss the experiments. Finally, we conclude the paper in Section 5.

\section{Related Work}

There are a variety of methods for lip-reading. These methods can fall into two categories: word-level and sentence-level lip-reading. In word-level methods, lip-reading is a classification task, whereas in sentence-level methods, sequence prediction is the problem. There are many pre-deep learning methods, which you can review by referring to \cite{zhou2014review, lombardi2013survey, mathulaprangsan2015survey}. In one of these methods \cite{bear2016decoding}, modeling is performed at the viseme-level using HMM (Hidden Markov model). After obtaining visemes, another phoneme HMM is trained for converting each viseme to a specific phoneme. This method is a non-deep approach and examined on little data.

\begin{figure*}[t]
	\centering
	\includegraphics[width=0.9\textwidth]{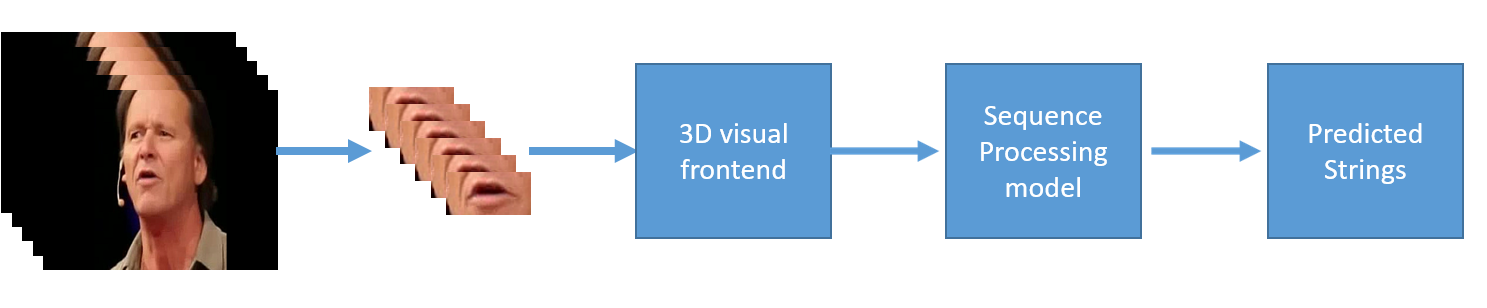}
	\caption{Traditional sequence to sequence methods.}
	\label{fig:fig1}
\end{figure*}

\begin{figure*}[t]
	\centering
	\includegraphics[width=0.9\textwidth]{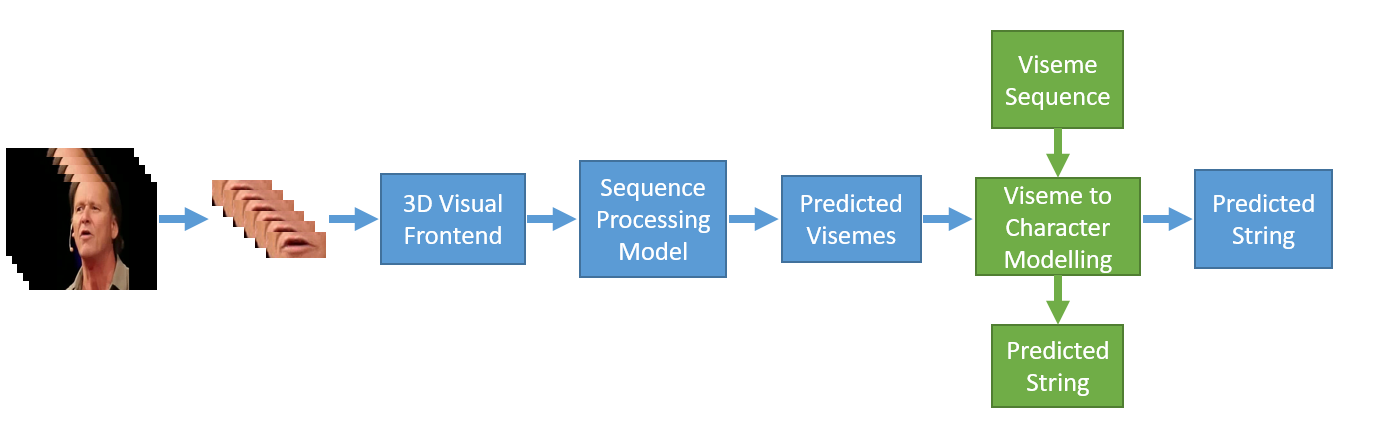}
	\caption{Our proposed lipreading method using external viseme to character model.}
	\label{fig:fig2}
\end{figure*}

The first method proposed for sentence-based lip-reading based on deep learning was called LipNet \cite{assael2016lipnet}. The LipNet architecture has 3 layers of STCNN (Spatio-Temporal CNN) followed by 2 Bi-GRUs (Bidirectional Gated Recurrent Unit). As an end-to-end model, LipNet is trained with CTC loss \cite{graves2006connectionist}. This method has been tested on GRID dataset. In the next method called WAS (Watch, Attend and Spell), the attention mechanism is used, and lip-reading is performed on LRS2 data, which is a real-world data \cite{chung2017lip}. This model is based on LAS (Listen, Attend and Spell) which has been developed for speech recognition task \cite{chan2016listen}.

The deep learning architectures are compared in \cite{afouras2018deep}. In this comparison, three new neural network architectures (Fully convolutional, Bidirectional LSTM, and Transformer \cite{vaswani2017attention}) are compared and the best performing network with respect to word error rate is the Transformer with a 50\% of WER (Word Error Rate) on LRS2. The fully convolutional network also has the best training and inference time.

In another recent work, an effective strategy for training lip-reading model has been proposed that uses speech recognition directly \cite{afouras2020asr}. This method, which is based on knowledge distillation, does not require manually annotated lip-reading data and the videos are unlabeled. This method predicts the speech in sentence-level and obtains state-of-the-art results on the LRS2 and LRS3 datasets.

In still another work, the main focus is on multilingual synergized lip-reading \cite{luo2020synchronous}. In this method, a model with higher accuracy in both languages can be achieved using data from two different languages. The main idea of this work is based on the fact that common patterns in lip movement exist in different languages because human vocal organs are the same. This method obtains state-of-the-art performance on the two challenging word-level lip-reading benchmarks, namely LRW (English) and LRW-1000 (Mandarin).

The authors proposed in \cite{martinez2020lipreading} a variable-length augmentation and used Temporal Convolutional Networks, suggesting another improvement to word-level lip-reading.

\section{Proposed method}

In this section, we propose a method in which external textual data can improve the lip-reading model accuracy. This section consists of two parts. In the first part, we will describe the highest accuracy that can be achieved in word-level lip-reading, and in the second part, we will explain our proposed method.

\subsection{Word-level lower bound error for greedy algorithm}

First, we used the available lip-reading text data to find the lower bound error of word-level viseme-to-character modeling. In this case, we first find text data vocabularies and the percentage of repetition of each word. Then, using the pronunciation list of the words and one of the suggested phoneme-to-viseme mapping \cite{harte2015tcd}, the viseme sequence for each word is obtained. Obviously, some words have the same viseme sequence (like art and heart). We then categorize these words and use a greedy algorithm to get the minimum error. The best choice for any viseme sequence, if there is more than one word, is the word that is repeated the most. In the experiment we performed on the LRS3 dataset \cite{afouras2018LRS3}, the lowest WER was 24.29\%. In addition, we tested this experiment on the LRS2 dataset, and the lowest WER was 27.16\%. But this is for the case that the context is not taken into account. In the following, we propose a method that can be used to achieve higher accuracy for viseme decoding.

\begin{table*}[th]
	\caption{Examples of viseme decoding results.}
	\centering
	\setlength\tabcolsep{15pt}
	\begin{tabular}{l | l}
		\toprule
		\midrule
		 Ground Truth                   & Predicted \\
		\midrule
		AND SO THIS IS WHAT I DID       & AND SO THIS IS WHAT I DID \\
		SO I SHOULD TALK ABOUT ART      & SO I SHOULD TALK ABOUT \textcolor{red}{YOU} \\
		WELL THE RANGE IS QUITE A BIT   & WELL \textcolor{red}{THERE HAD CHASE} QUITE A BIT \\
		THIS IS NELL REMMEL             & THIS IS \textcolor{red}{THE TROUBLE} \\
		NEXT IS SYLVIA SLATER           & NEXT IS \textcolor{red}{SILVER IS NOT HER} \\
		THAT IS MY MOM AND DAD          & THAT IS MY \textcolor{red}{PEOPLE THEN} \\
		\midrule
        \bottomrule
	\end{tabular}
	\label{tab:examples}
\end{table*}

\subsection{Lip reading using external viseme decoding}

In recent years, with the advances in the field of deep learning, significant progress has been made in many computer vision problems. One of the most difficult tasks in this field is lip reading and viseme decoding. There are usually small datasets available for this task in a variety of languages. Also, providing data in this area for the available methods is very costly. Because these methods require curriculum learning, word-level annotation is needed.

In this paper, we intend to solve this problem separately using available sequence-to-sequence methods. This allows us to use raw textual data in a language directly for viseme decoding. With respect to the lower bound error mentioned in the previous section, we expect the inaccuracy of this model to be much lower because in this task the context is considered and each word is not decoded separately.

The lip-reading methods mentioned in the previous section perform modeling at the sentence level, and with respect to the main challenge in lip-reading, which is viseme decoding, it is expected that the video-to-viseme conversion will be done with greater accuracy. Also, lip-reading can be done at the sentence level using these two proposed models.

In fact, both sub-models have their own advantages which improve accuracy. We describe these two models in order of use. In the first model, which aims to convert video to viseme, the existing methods can be used exactly and we do not need more data or any change in the structure of the network. Nevertheless, we expect to achieve higher accuracy due to the smaller number of classes. Moreover, there is no need for a lip-reading dataset in the second model. To train this model, we need raw textual data in the target language. Also, training data can be obtained as needed by having a phoneme sequence for each word and using the phoneme-to-viseme mapping. The only challenge when constructing this data set is to obtain the sequence of words for an utterance. There are several solutions to this problem called G2P (grapheme-to-phoneme).

As shown in Figure~\ref{fig:fig1}, in traditional sequence-to-sequence methods, the first step is determining the mouth area using the facial landmarks for cropping the ROI (region of interest). This sequence is then modeled using a 3D visual front-end (usually using 3D-CNN) followed by a sequence processing model. In fact, in these methods, the feature extraction of lip movements is obtained using a 3D convolutional network, and the output is a set of probabilities for each character. But in the proposed method, shown in Figure~\ref{fig:fig2}, another network is trained with independent data. In fact, in this method, two networks are trained, one of which converts video to viseme and the other predicts the characters using visemes sequence.

\section{Experiments}

In this section, we will refer to the experiments we performed as well as the results. The experiments are divided into two parts. In the first part, we describe the conversion of viseme to character. We also compare the accuracy obtained for the case of using raw data or existing data for lip-reading. Also, in the second part, we perform lip-reading at the character level using the obtained model, along with a viseme level lip reading model.

Note that our goal was not to achieve the best reported results, but due to lack of time, we intended to show that the proposed method can improve the baseline. We believe that this improvement can be achieved by any other method. In the future, we will try to first replicate the results reported in~\cite{chung2017lip} and then incorporate the proposed method into them to make a better comparison with the state-of-the-art results.

\subsection{Viseme decoding}

As explained earlier, usually there is little data for training a lip-reading model. In this experiment, we want to show how increasing the unlabeled text data can affect the accuracy of the viseme decoding. In this step, we first trained the model using the textual data of the LRS2 dataset. Given that the size of the LRS2 samples is not large in terms of textual data, in the second case we used OpenSubtitles corpus \cite{tiedemann2004opus} for viseme to character modeling and measured the effect of this improvement. 

We selected 6 million samples from the OpenSubtitles corpus for this purpose. But since only the word sequence is available in this corpus, in the first step we used CMU Pronouncing Dictionary to convert this word sequence into a phoneme sequence. In this step, we also removed the sentences containing words that were not in the dictionary. Given the fact that the main purpose here is a kind of language modeling, (i.e. the possibility of occurrence of different consecutive visemes is important to us), we are not very sensitive to choose an accurate transcription for all the words. Consequently, we used a simple method for this purpose. In this way, we only select the first transcript for each word in the dictionary with multiple transcripts. We used the CMU dictionary to convert the word sequence in the OpenSubtitles data into a phoneme sequence. Subsequently, the viseme sequence for each sample is obtained using the phoneme to viseme mapping. After preparing these data, the model was trained and some of the results are shown in Table~\ref{tab:examples}. As mentioned above, we used a sequence-to-sequence network with a two-layer GRU with a cell size of 1024. We also used attention mechanism \cite{chorowski2015attention} in order to achieve a better result.

\begin{table}[t]
	\caption{WER and CER for Viseme to Character Modeling.}
	\centering
	\setlength\tabcolsep{12pt}
	\begin{tabular}{l | c  c}
		\toprule
		\midrule
		Dataset             & CER               & WER  \\
		\midrule
		LRS2                & 26 \%             & 37\%     \\
		OpenSubtitles       & \textbf {10 \%}   & \textbf{16 \%} \\     
		\midrule
		\bottomrule
	\end{tabular}
	\label{tab:table1}
\end{table}

The results of Table~\ref{tab:table2} show that using the OpenSubtitles corpus reduced the relative CER (Character Error Rate) by approximately 62\%, and the relative WER error by around 57\%. The results of this experiment indicate that by having more training data for viseme decoding, a better language model can be obtained, and this will improve the accuracy of this decoding. Of course, there is definitely an upper bound for this improvement, and it cannot be claimed that by increasing the data, the error can be reduced to zero. But it shows that it is easier to improve the accuracy with that as there is a lot of unlabeled textual data in any language.

\subsection{Sentence-level lip-reading}

In this step, using the proposed models for character-level lip-reading, we perform viseme-level lip-reading. For this purpose, we had to first train the video-to-viseme model, for which we need to have a sequence of viseme for each video sample. Here again, we used a dictionary similar to the previous experiment. After the phoneme sequence for each training sample was obtained, the phoneme-to-viseme mapping was used to convert the phoneme sequence into the viseme sequence. After that, the video-to-viseme model was trained using a simple sequence to sequence model. The results of this experiment are shown in Table~\ref{tab:table2}.

\begin{table}[t]
	\caption{Performance on LRS2 dataset.}
	\centering
	\setlength\tabcolsep{12pt}
	\begin{tabular}{l | c  c}
		\toprule
		\midrule
		Method              & WER               & CER  \\
		\midrule
		Video to Viseme     & 62.3 \%            & 33.9 \%     \\
		\midrule
		WAS~\cite{chung2017lip}       & 73.9 \%            & 49.9 \%     \\
		Proposed method     & \textbf {69.5 \%}  & \textbf{46.1 \%}  \\
		\midrule
		\bottomrule
	\end{tabular}
	\label{tab:table2}
\end{table}

In the first row of this table, the video-to-viseme result shows that this model is able to do this task with an acceptable degree of accuracy even by using a simple model. So far, we have only viseme sequence as output. In the second part of the table, the result of combining this model with the model prepared in the previous step is provided, along with a comparison with the result obtained through the method used in~\cite{chung2017lip}. Considering that a simple architecture is used in both trained models, the improvement of accuracy compared to~\cite{chung2017lip} is considerable.

As shown in Table~\ref{tab:table2}, in the case of viseme-level modeling, the character error rate is 33.9\%, which is 16\% more accurate than in the case of word-level modeling (i.e. the second row). Also, when external textual data is used for viseme decoding, we achieve higher accuracy than if the network implicitly learns the language model probabilities.

\section{Conclusion}

Lip reading is one of the most challenging tasks in the field of computer vision. There is, in fact, scant data available on this task for many languages. We introduced a new method to use external text data for lip-reading. We can achieve higher accuracy for lip-reading by utilizing the raw text data of a specific language, a grapheme-to-phoneme as well as a viseme mapping. The experimental results indicated that the proposed method can improve the accuracy of viseme decoding and outperforms the case where only lip-reading text data is used for language modeling by a wide margin. We also incorporated this model into the viseme-level model for lip-reading and achieved higher accuracy than the case where only video data was used for training. One of the limitations of our work is the use of the same phoneme sequence for words with more than one correct pronunciation. These words can be pronounced in several ways and we do not know which pronunciation is used in the video. To get better results, the output of an automatic speech recognition system can be considered as future work. Furthermore, due to internal limitations, we had to use a simple sequence-to-sequence model for both tasks. Therefore, as another future work, we will incorporate our proposed method into state-of-the-art systems to show how it can improve the overall performance of a lip-reading system.

\section{Acknowledgement}

The authors would like to extend their gratitude to the Speech Laboratory of the Brno University of Technology for providing access to computational servers and LRS datasets.

\bibliographystyle{IEEEtran}

\bibliography{mybib}

\end{document}